\documentclass[runningheads]{llncs}
\usepackage[T1]{fontenc}
\usepackage{graphicx}
\usepackage{orcidlink}
\usepackage{bbding}  
\usepackage{enumitem} 
\usepackage{multirow}
\usepackage{multicol}
\usepackage{makecell}
\usepackage{amssymb}
\usepackage{listings}
\usepackage{xcolor}
\usepackage{amsmath}
\usepackage{pgf-umlsd}
\usepackage{tikz}
\usetikzlibrary{positioning,arrows.meta}
\usepackage{subcaption}

\definecolor{jsonkey}{rgb}{0.0,0.3,0.6}
\definecolor{jsonstring}{rgb}{0.6,0.1,0.1}
\definecolor{jsonbg}{rgb}{1, 1, 1}
\definecolor{jsonnum}{rgb}{0.1,0.5,0.1}

\lstdefinestyle{jsonstyle}{
    backgroundcolor=\color{jsonbg},
    basicstyle=\ttfamily\footnotesize,
    breaklines=true,
    showstringspaces=false,
    numbers=left,
    numberstyle=\tiny\color{gray},
    stringstyle=\color{jsonstring},
    keywordstyle=\color{jsonkey},
    commentstyle=\color{gray}
}

\hypersetup{
  colorlinks=true,
  linkcolor=black,
  citecolor=black,
  urlcolor=black
}

\begin{document}
\title{Toward AI-Augmented Cooperative Engineering Workflows: Requirements and Architecture from the European Rover Challenge}

\author{Ahmed R. Sadik\inst{1}{\Envelope}\,\orcidlink{0000-0001-8291-2211} \and
Frank Joublin\inst{1}\orcidlink{0000-0002-4421-1737} \and
Mariusz Bujny\inst{2}\orcidlink{0000-0003-4058-3784} \and
Antonello Ceravola\inst{1}\orcidlink{0000-0002-1075-459X} \and
Joan Smith\inst{3}
} 
\authorrunning{Sadik et al.}
\institute{Honda Research Institute Europe, Offenbach, 63073, Germany
\email{\{ahmed.sadik, frank.joublin, antonello.ceravola\}@honda-ri.de}
\and
NUMETO, Katowice, Poland
\email{mariusz.bujny@numeto.eu}\\ \and
Honda Research Institute USA, Ohio, USA
\email{jrsmith@honda-ri.com}}
\maketitle

\begin{abstract}
The growing availability of Artificial Intelligence (AI) tools creates new opportunities to support engineering design processes, yet their current use often remains limited to isolated tasks such as coding, documentation, or information retrieval. Less attention has been given to how AI can support cooperative engineering workflows at the process level, where teams must coordinate requirements, tasks, communication, knowledge transfer, and subsystem integration. This paper investigates this challenge in the context of the European Rover Challenge (ERC), where student teams design and integrate complex rover systems within a single academic cycle under strict time constraints and high subsystem interdependence. We conducted a role-adaptive 40-question survey with ERC-2025 teams, yielding 104 responses from 14 teams. The survey examined team structure, knowledge transfer, task management, integration practices, communication patterns, and current AI usage. The results reveal recurring workflow bottlenecks, including limited documentation, unclear requirements, fragmented communication, informal task monitoring, and substantial integration rework. Based on these findings, we derive requirements for AI-augmented cooperative engineering workflows and propose an initial assistant-system architecture that connects user-facing interfaces, credential management, service selection, specialized AI services, and external engineering tools. The proposed architecture aims to support task clarification, requirement and compliance management, communication summarization, integration-risk detection, and continuous knowledge capture. In doing so, the paper contributes empirical requirements and an architectural direction for AI-augmented cooperative engineering workflows in hybrid human--AI team settings.


\end{abstract}

\section{Introduction}
\label{sec:introduction}

Engineering design in complex, time-constrained projects is not only a technical endeavor, but also a cooperative, organizational, and psychological one. Such projects require teams to coordinate people, tasks, requirements, communication channels, engineering artifacts, and interdependent technical subsystems over time. Organizational psychology has long shown that team performance, quality outcomes, and sustainability of effort depend on how work is structured, distributed, communicated, and regulated over time \cite{hackman1975development,salas2015understanding}. In project-based environments, poorly specified tasks, fragmented communication, and informal coordination mechanisms are strongly associated with cognitive overload, stress accumulation, and increased rework ultimately leading to inefficiency and burnout \cite{quick1990healthy,demerouti2001job}. From the perspective of cooperative information systems, these problems are not only social or managerial issues, but also symptoms of weak information infrastructures for supporting shared awareness, task traceability, knowledge continuity, and cross-subsystem coordination \cite{bravo2025creation}.

The European Rover Challenge (ERC) \cite{ercWebsite} teams mission is to design a  unique rover and drone \cite{sadik2023self}, which can mimic real life space tasks. ERC  provides a particularly revealing context in which to study these dynamics. ERC teams operate under intense temporal pressure, strict external constraints, and high interdependence between subsystems, while simultaneously coping with high turnover and uneven experience levels. From an organizational psychology perspective, such conditions closely resemble what are described as high-demand, low-formalization systems, in which workload regulation is largely left to individuals rather than supported by structural mechanisms \cite{ilgen2005teams}. While these environments can foster learning and motivation, they also systematically increase the risk of role ambiguity, hidden over-work, coordination breakdowns, and late-stage crisis management. At the same time, they provide a useful empirical setting for studying how cooperative information systems could support engineering teams by connecting requirements, tasks, communication, documentation, and integration activities more continuously.

Research on work design and Job Demands--Resources (JD-R) theory suggests that when demands such as time pressure, task uncertainty, and coordination overhead are not balanced by adequate resources clear goals, feedback, autonomy with support, and cognitive offloading tools performance and well-being degrade simultaneously \cite{bakker2023job,yamauchi2013robotic}. In engineering teams, this imbalance often manifests as excessive rework, knowledge loss, uneven workload distribution, and reliance on informal heroics during integration and testing phases. Importantly, these effects are not merely individual failures but systemic outcomes of how work is organized and supported. This makes the design of cooperative information systems especially relevant: better support for task clarification, knowledge capture, dependency tracking, and shared situational awareness can function as structural resources that reduce coordination friction and improve workflow reliability.

Against this backdrop, this paper investigates how AI could support cooperative engineering workflows by analyzing organizational and workflow practices within ERC teams and by deriving requirements for an AI-augmented assistant architecture. Using a role-adaptive survey completed by 104 participants across 14 ERC-2025 teams, we examine team structures, task allocation and monitoring practices, communication patterns, knowledge transfer mechanisms, integration challenges, and current AI usage. By linking observed bottlenecks to established theories of work design and team functioning, the study identifies leverage points where AI particularly low-intrusion, context-aware tools could reduce coordination overhead, clarify work, and alleviate excessive workload without disrupting the autonomy and creativity central to student engineering teams.

The paper makes three contributions. First, it provides empirical insight into recurring cooperation bottlenecks in ERC engineering workflows, including unclear requirements, fragmented communication, limited documentation, informal task monitoring, and late-stage integration rework. Second, it translates these findings into requirements for AI-augmented cooperative engineering workflows, covering requirement and compliance management, task clarity, workload awareness, communication summarization, knowledge capture, and integration-risk detection. Third, it proposes an initial architecture for an AI-augmented cooperative information system that connects user-facing interfaces, credential management, service selection, specialized AI services, and external engineering tools to support hybrid human--AI engineering teams across design, development, integration, and testing.

\section{Survey Structure and Question Flow}

\begin{figure}
  \centering
  \includegraphics[width=0.8\linewidth]{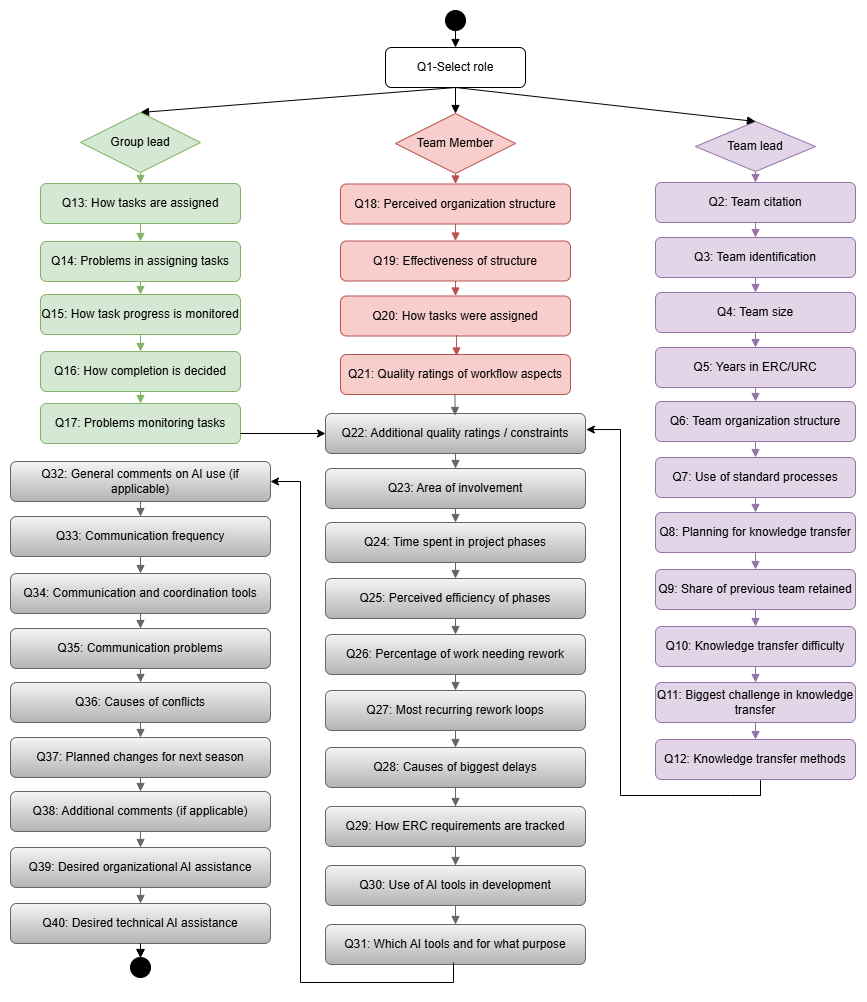}
  \caption{Role-adaptive branching structure of the ERC survey.}
  \label{fig:survey-flow}
\end{figure}

The survey was designed to capture multiple perspectives on organizational and workflow practices within ERC teams, with the aim of identifying where AI-based support could most effectively improve cooperative engineering processes and inform the requirements of an AI-augmented assistant architecture (see Fig.~\ref{fig:survey-flow}). To balance analytical depth with respondent burden, the survey employed a role-adaptive branching structure. Participants first identified their role within the team as team leader, group leader, or team member, after which they were guided to role-specific question blocks.

Team leaders were asked about overall team structure, experience levels, and mechanisms for knowledge transfer. Group leaders focused on task assignment, coordination, and monitoring within and across subsystems, while team members reported on day-to-day execution practices and workflow realities. Following these role-specific sections, all participants completed a shared block of questions addressing core workflow elements, including task definitions, dependencies, deadlines, integration planning, communication practices, and current AI usage. This structure enabled the survey to link organizational choices and coordination practices to concrete outcomes such as rework, delays, and perceived inefficiencies. By combining role-specific insights with a common workflow perspective, the design supports a targeted analysis of where AI-driven tools could assist with task clarification, coordination, knowledge transfer, requirement tracking, and integration across ERC teams. These insights provide the empirical basis for deriving design requirements and motivating the architecture proposed later in the paper.

\section{Empirical Findings and Derived Requirements}

This section presents the empirical findings obtained from the survey responses and interprets them as input for deriving requirements for AI-augmented cooperative engineering workflows. The responses, along with the full list of questions, are published as an anonymized dataset \cite{sadik_dataset_2026} to ensure reproducibility. In order to provide a coherent interpretation of the survey, the results are organized into thematic categories that reflect key dimensions of team operations. Each thematic category is presented as a separate subsection and concludes by identifying implications for AI-supported cooperative information systems.

\subsection{Team Structure and Organizational Model}

The survey responses represent a broad cross-section of roles within ERC teams. Participants included regular team members (57.7\%), group leaders responsible for individual subsystems (28.8\%), and overall team leaders (13.5\%). Teams varied substantially in size, ranging from approximately ten to more than forty active members, with most reporting around thirty participants. Experience levels likewise differed widely, from teams participating in ERC for the first time to institutions with more than a decade of involvement in ERC or the related University Rover Challenge (URC)~\cite{university_rover_challenge}.

Across teams, organizational structures were predominantly functional, with members grouped by subsystem such as mechanical, electrical, or software engineering. This model was reported by the majority of respondents (70\%), while mixed (15\%) and matrix-style (13\%) structures were considerably less common. Respondents’ domain involvement reflected the multidisciplinary nature of rover development, with frequent participation in robotic arm design (46\%), software (44\%), mechanical systems (42\%), and management activities (37\%), alongside electrical systems (31\%), chassis design (23\%), navigation (21\%), and telecommunications (20\%). This distribution underscores the high degree of interdependence between subsystems, particularly during integration and testing phases.

Despite this diversity, participants expressed moderate confidence in the effectiveness of their team structures. On a five-point scale, just over half of respondents rated their team organization as effective (score of 4), while a quarter selected a neutral score (3). overall, the findings depict teams that are technically diverse and predominantly organized around functional subsystems, yet vulnerable to coordination challenges across subsystem boundaries. These structural characteristics are consistent with later findings on integration delays and rework, and they motivate the need for AI-supported cooperative information systems that can monitor cross-subsystem dependencies, anticipate coordination bottlenecks, and support more balanced workload distribution across teams.

\subsection{Knowledge Transfer and Documentation}

\begin{figure}[!ht]
    \centering
    \includegraphics[width=0.6\linewidth]{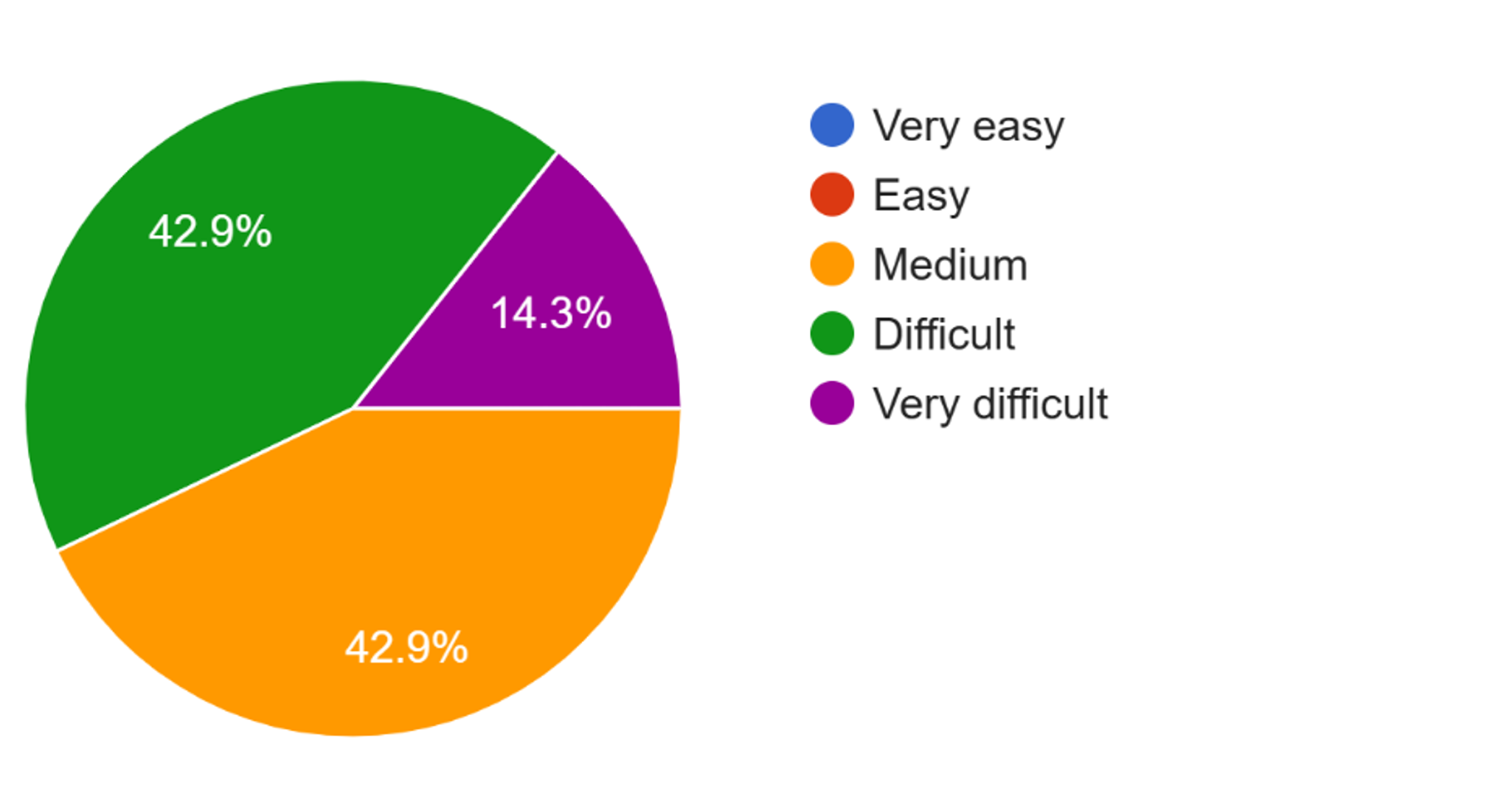}
    \caption{Self-reported difficulty of knowledge transfer, showing that all teams rated the process as Medium, Difficult, or Very Difficult.}
    \label{fig:kt-difficulty}
\end{figure}

Knowledge transfer has become one of the most challenging aspects of team operations \cite{nokes2009mechanisms}. Retention rates (Q9) varied widely: 14.3\% of teams retained fewer than 10\% of their 2024 members, while 35.7\% retained more than half, resulting in substantial variation in institutional continuity. Consistent with this, respondents in Q10 rated knowledge transfer as a significant challenge. As shown in Fig.~\ref{fig:kt-difficulty}, no team described the process as ``Very easy'' or ``Easy''; instead, 42.9\% rated it as ``Medium'' and another 42.9\% as ``Difficult,'' with a smaller share selecting ``Very difficult.'' Most teams relied primarily on documentation for handover (92.9\%), complemented by coaching or mentoring (78.6\%) and hands-on sessions (64.3\%), as reported in Q12. Workshops were less common, and no team used recorded sessions, indicating limited production of reusable on-boarding materials.

For project management (Q7 and Q8), some teams adopted lightweight organizational tools such as Kanban boards, agile cycles, or hybrid Waterfall--Agile workflows. These practices were supported by tools such as Discord, Notion, or GitHub. More mature teams reported coding standards and structured life-cycle models, such as the V-model or INCOSE-inspired processes. This variability contributes to uneven documentation quality and inconsistent integration readiness. Most teams attempted proactive knowledge-transfer planning such as on-boarding workshops, pairing juniors with seniors, or maintaining cloud-based repositories. However, long-term continuity remained limited because subsystem implementations evolved faster than the corresponding documentation. Teams frequently noted that critical design rationale was missing, requiring newcomers to reverse-engineer previous work.

Finally, Q11 highlighted the most common challenges: limited time to produce high-quality documentation, lack of foundational knowledge among new members, and difficulties handling large mechanical files under version control. These issues motivate the need for AI-supported cooperative information systems that can continuously generate and update documentation, capture design rationale, assist on-boarding, and extract institutional knowledge from communication logs, design files, and version-control systems. In the context of the architecture proposed later in this paper, knowledge transfer therefore becomes not only a documentation problem, but also a requirement for continuous knowledge capture and retrieval across distributed tools, teams, and engineering artifacts.

\subsection{Task Assignment, Monitoring, and Execution}

\begin{figure}[!ht]
    \centering
    \includegraphics[width=0.8\linewidth]{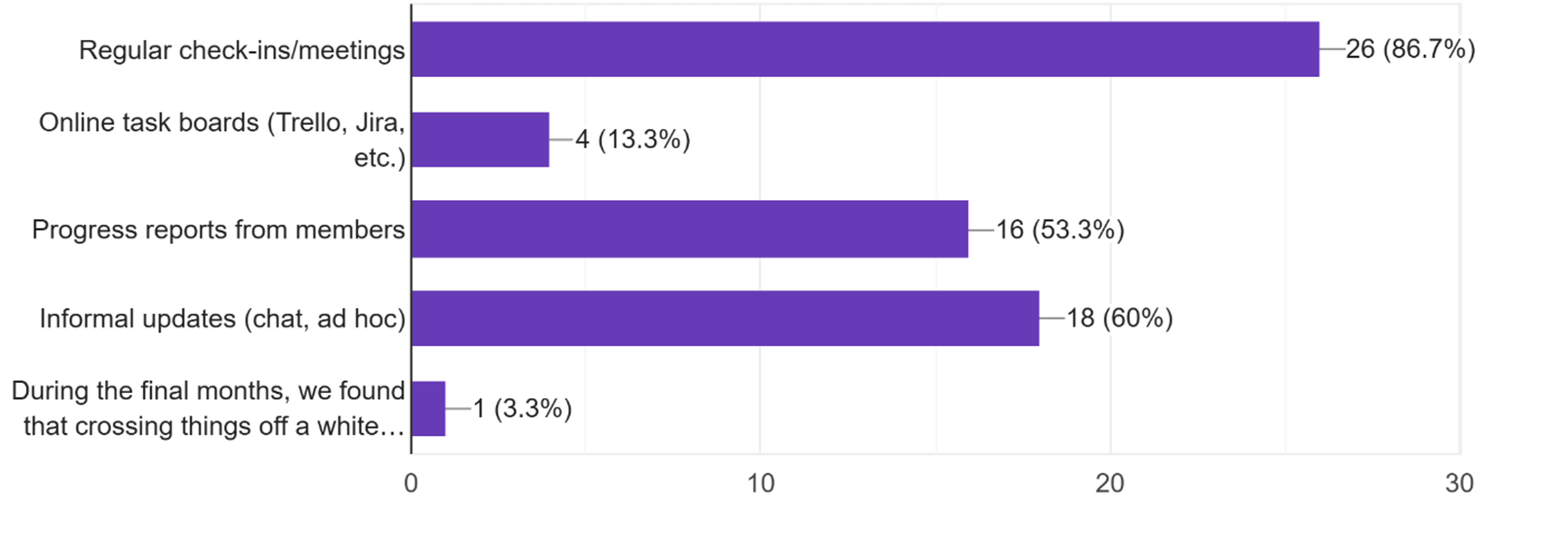}
    \caption{Methods used to monitor task progress, illustrating the predominance of informal updates and limited use of structured tracking tools.}
    \label{fig:task-monitoring}
\end{figure}

The assignment of tasks within ERC teams is highly dependent on informal communication \cite{head2009systems}. According to Q13 and Q20, most group leaders (90\%) assign tasks verbally during meetings and/or through chat platforms such as Discord (70\%), while only a minority consistently use structured project-management tools (3\%). From the members' perspective, tasks are usually obtained through group discussions or directly from group leaders, with a significant proportion selecting tasks by themselves (30\%) or working on tasks that were never formally assigned (23\%). These patterns, reflected in Q14, indicate diffuse responsibility, unclear task ownership, and limited traceability by at least 23\% of the group leader responses. Monitoring practices exhibit similar informality. As shown in Fig.~\ref{fig:task-monitoring} and reported in Q15, progress is most often tracked through regular or informal updates, and only a small fraction of respondents use online task boards (13\%) for systematic monitoring. This lack of structured oversight hinders early detection of delays.

The qualitative responses of Q16 and Q17 are summarized in the following two tables. Missing or delayed updates were especially problematic in tightly coupled subsystems, where a single stalled task could cause cascading delays across team groups. In general, this category reveals structurally similar challenges in task assignment and monitoring, such as informal communication, unclear requirements, inconsistent update practices, and incomplete validation procedures. These limitations reduce workflow predictability and contribute to late-stage rework and integration issues.

\begin{table}[h!]
\centering
\footnotesize
\begin{tabular}{p{10.2cm} c}
\textbf{Assigning Tasks - Major Problems} & \textbf{\%}\\
\hline
Unclear task definitions/requirements & 23\% \\
Ineffective or unused task mgmt tools; poor overview & 17\% \\
Member availability / scheduling constraints & 13\% \\
Time pressure; insufficient time or people & 13\% \\
Planning, prioritization, dependencies, coordination & 13\% \\
Low engagement, absenteeism, unreliable follow-through & 13\% \\
Uneven skills/experience; missing specialization & 7\% \\
\end{tabular}
\end{table}

\begin{table}[h!]
\centering
\footnotesize
\begin{tabular}{p{10.2cm} c}
\textbf{Monitoring Tasks - Major Problems} & \textbf{\%}\\
\hline
Inconsistent/tardy progress updates; tools not adopted & 17\%  \\
Communication issues (absenteeism, no reporting, no help-seeking) & 17\% \\
Schedule/deadline mgmt problems; interdependencies; crunch time & 17\% \\
Monitoring workload/verification overhead & 13\% \\
Lack of Definition of Done; unclear criteria; poor granularity & 13\% \\
Delays and low commitment/accountability & 13\% \\
Scope/quality expectation misalignment & 3\% \\
Tracking many parallel tasks; coordination overhead & 3\% \\
\end{tabular}
\end{table}

These findings motivate the need for AI-supported cooperative information systems that can assist with task clarification, task decomposition, ownership tracking, progress monitoring, and dependency awareness. In the architecture proposed later in the paper, this corresponds to services that infer task status from communication and version-control activity, identify missing updates or unclear acceptance criteria, and provide proactive risk or delay notifications before stalled work propagates across tightly coupled subsystems.

\subsection{Workflow Quality, Rework, and Delays}

Across teams, workflow quality showed substantial variability, particularly in upstream elements such as task definitions, acceptance criteria, and dependency specification \cite{deschamps2015impact}. According to Q21, these components were frequently rated as only ``sufficient'' or ``insufficient,'' indicating unclear requirements and inconsistent planning for up to 25\% of the team members. Such challenges propagate downstream, creating misaligned expectations and avoidable rework. The time allocations reported across the project phases (Q24) varied widely, especially for recruitment, concept definition, and documentation, where durations ranged from less than two weeks to more than eight. Teams also indicated in Q25 that many phases - most notably testing, final validation, and documentation could have been executed more efficiently, reflecting systemic issues with early integration.

\begin{figure}[!ht]
    \centering
    \includegraphics[width=\linewidth]{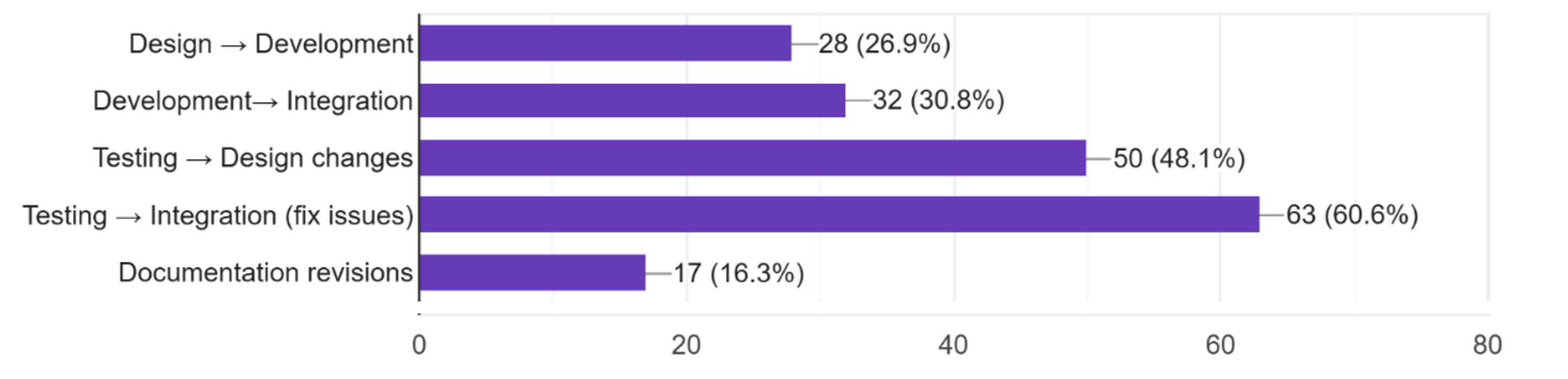}
    \caption{Most frequent rework loops reported by teams, highlighting instability at key transitions such as testing--integration and testing--design.}
    \label{fig:rework-loops}
\end{figure}

Rework emerged as one of the most significant challenges identified in the survey. According to Q26, many teams estimated that 20--40\% of their time in design, development, integration, and testing was spent revisiting previous work. Documentation was particularly problematic, with several teams reporting more than 50\% rework due to outdated or incomplete material. The most frequent rework cycles occurred between the testing phase and the subsequent integration phase (60.6\%), followed by rework that returned from testing back to the design phase (48.1\%). These patterns indicate that incompatibilities between subsystems and incomplete or flawed assumptions were often discovered late in the development process (Fig.~\ref{fig:rework-loops}).

\begin{figure}[!ht]
    \centering
    \includegraphics[width=\linewidth]{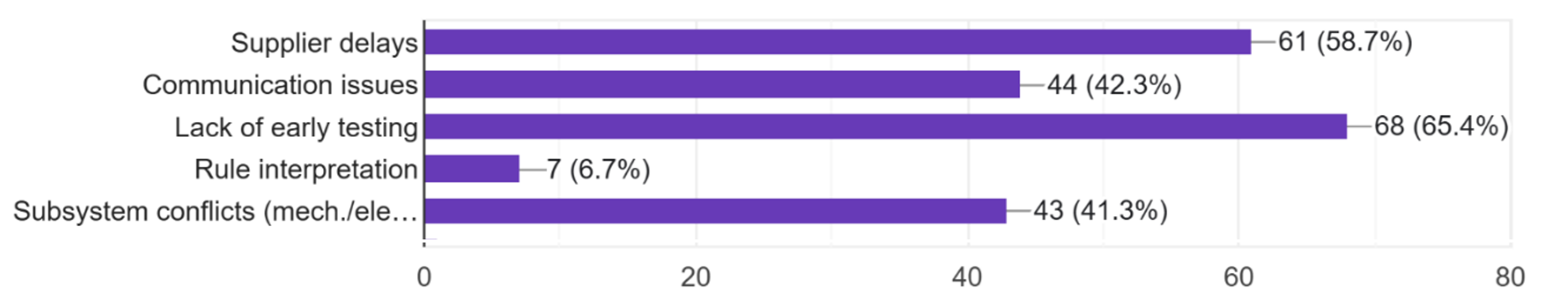}
    \caption{Primary causes of delays reported by respondents, demonstrating the impact of insufficient early testing, supplier issues, and communication gaps.}
    \label{fig:delays}
\end{figure}

The delay patterns reported in Q28 showed similar underlying causes. The most frequently cited issues were insufficient early testing (65.4\%), supplier delays (58.7\%), communication problems (42.3\%), and conflicts between subsystems (41.3\%). Additional but less common contributors included unclear rule interpretation and institutional constraints such as limited funding or administrative delays (Fig.~\ref{fig:delays}). Together, these results illustrate that late identification of issues - whether technical, organizational, or logistical remains a recurring and costly characteristic of ERC workflows.

These findings motivate the need for AI-supported cooperative information systems that can improve early-phase planning, clarify requirements, and coordinate subsystem workflows before integration problems become costly. In the architecture proposed later in the paper, this corresponds to services for dependency analysis, interface checking, integration-risk prediction, and continuous monitoring of communication logs, design histories, and version-control data. Such support would allow teams to detect early conflicts, identify unstable assumptions, and reduce rework across the design, development, integration, and testing phases.

\subsection{ERC Requirement Tracking and Communication}

Requirement tracking in ERC teams is largely manual and decentralized. According to Q29, most teams extract rules into spreadsheets maintained in Google Sheets or Excel, updating compliance and responsibilities by hand. More structured groups use Notion databases, Jira or Trello boards, or simple traceability matrices derived from scoring sheets, but standardized or automated requirements-engineering workflows remain rare. The responsibility typically lies with a single lead, which makes the process vulnerable to omissions, delayed updates, and inconsistent interpretation. From the perspective of cooperative information systems, this indicates a weak connection between external rules, internal requirements, task responsibilities, and subsystem-level implementation work.

Communication patterns exhibit similar fragmentation. As reported in Q33, most respondents communicated weekly (67.3\%), while only about a third communicated daily and a notable minority interacted bi-weekly or monthly. Teams relied on a wide assortment of channels - WhatsApp (70.2\%), Discord (60.6\%), Google Docs (35.6\%), and Notion (32.7\%), as captured in Q34, often using them simultaneously without unified policies. This dispersion contributed to information loss and inconsistent alignment between subsystems.

\begin{figure}[!t]
    \centering
    \includegraphics[width=\linewidth]{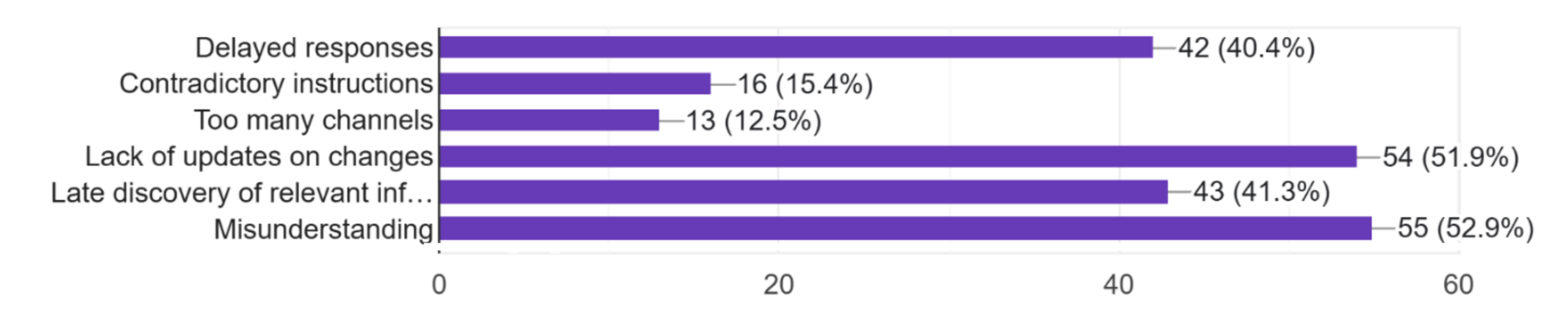}
    \caption{Most frequently reported communication problems, showing issues with misunderstandings, missing updates, and delayed responses.}
    \label{fig:comm-problems}
\end{figure}

The most frequent communication issues were misunderstandings (52.9\%), late or missing updates (51.9\%), delayed responses (40.4\%), and contradictory instructions (15.4\%), as reported in Q35 and shown in Fig.~\ref{fig:comm-problems}. Respondents also noted difficulty tracking critical information scattered across chats, documents, and platform-specific threads. These gaps often led to late discovery of interface changes, outdated assumptions, and duplicate effort. The conflicts arising from these issues were dominated by integration mismatches and scheduling conflicts (both 42.3\%), unclear responsibilities (33.7\%), and uneven availability of members (36.5\%), as indicated in Q36. Several respondents also described interpersonal friction under deadline pressure, particularly when communication breakdowns accumulated late in the season.

These findings highlight the absence of centralized requirement management and coherent communication architectures. The resulting fragmentation directly contributes to integration delays, rework, and misaligned subsystem development. It also motivates the need for AI-supported cooperative information systems that can connect requirements, responsibilities, communication updates, and design decisions across tools and teams. In the architecture proposed later in the paper, this corresponds to services for automated requirement decomposition, compliance tracking, cross-channel communication summarization, contradiction detection, and continuous identification of design-impacting updates. Such support would help maintain shared situational awareness throughout the project and reduce the risk that critical requirement or interface changes remain hidden in isolated communication channels.

\subsection{Use of AI Tools and Desired AI Support}

AI tools are widely used across ERC teams, with 78.8\% of respondents reporting the use of systems such as ChatGPT, Copilot, or similar tools. Their use is primarily focused on execution-level technical tasks, including software development, CAD modeling \cite{sadik2025human}, debugging \cite{sadik2023coding}, documentation drafting, and interpretation of technical information. More specialized applications, such as topology optimization, PCB verification, or visual asset generation, are used by only a small subset of teams. Despite this broad technical adoption, respondents consistently reported that their greatest unmet needs lie at the organizational and system-engineering level. In particular, they expressed strong interest in AI support for task management, workload coordination, requirement handling, communication summarization, and maintaining shared situational awareness across teams.

The technical expectations expressed in Q40 aligned closely with the bottlenecks identified elsewhere in the survey. Respondents highlighted the need for assistance in detecting subsystem incompatibilities, diagnosing wiring or communication failures, identifying integration risks, and predicting the behavior of the rover in simulation under fault conditions. Frequently mentioned capabilities also included verification of mechanical and electrical tolerances, PCB design checking, interpretation of data-sheets, and automatic generation of test plans or validation procedures. These expectations reflect a desire for AI tools that extend beyond code assistance toward system-level diagnostic and reasoning capabilities. Although the adoption of AI among ERC teams is widespread, its use remains concentrated on local technical tasks rather than on cooperative workflow support. Q39--Q40 show that teams most strongly want AI tools that support project coordination, requirement traceability, integration planning, and subsystem compatibility analysis. These needs mirror the organizational and integration bottlenecks identified throughout the survey and motivate the development of AI-supported cooperative information systems that provide proactive, system-aware support throughout the rover development life cycle. In the architecture proposed later in the paper, this corresponds to a modular assistant system in which user-facing interfaces are connected to specialized AI services for requirement handling, task coordination, documentation support, communication summarization, and integration-risk analysis.

\section{Proposed AI-Augmented Cooperative Engineering Architecture}
\label{sec:architecture}

The empirical findings indicate that ERC teams do not primarily lack isolated AI tools; rather, they lack an integrated cooperative information system that connects requirements, tasks, communication, documentation, and engineering artifacts across the development workflow. The proposed architecture therefore aims to support AI-augmented cooperative engineering by acting as an intermediary layer between team members, specialized AI services, and external engineering tools. Instead of replacing existing tools or imposing a rigid project-management process, the architecture is intended to provide low-intrusion support for the activities identified in Section~3: task clarification, requirement tracking, knowledge capture, communication summarization, progress awareness, and integration-risk detection.

\vspace{-5mm}

\begin{figure}[!ht]
    \centering
    \includegraphics[width=\linewidth]{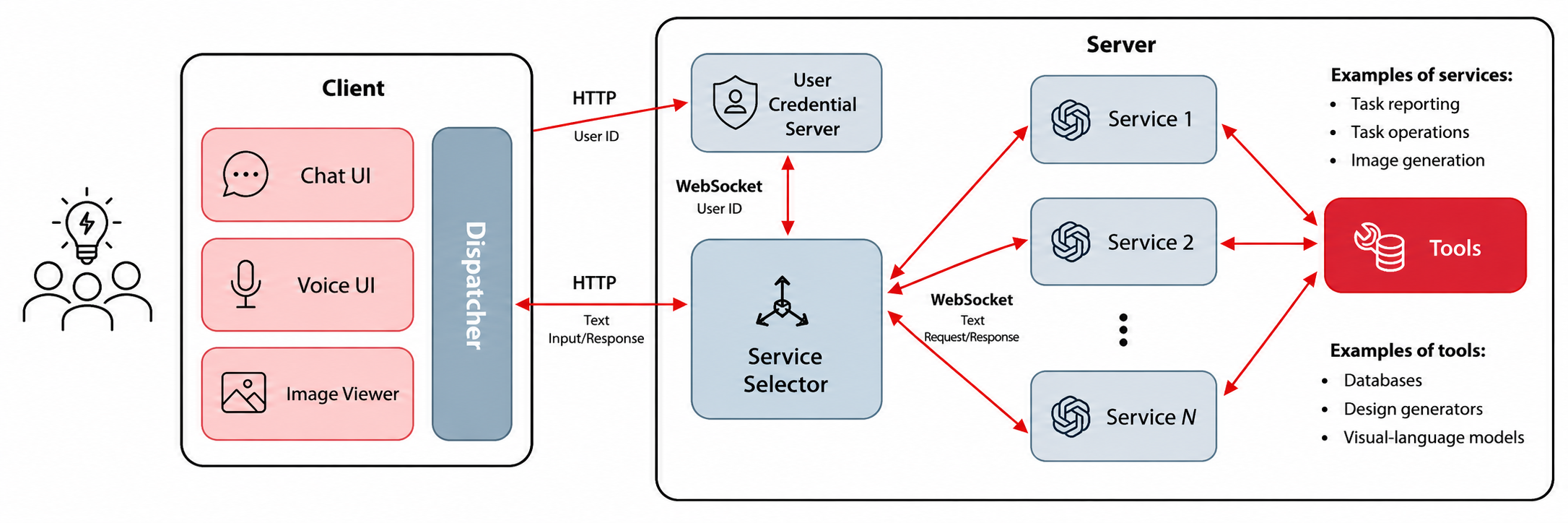}
    \caption{Proposed architecture of the AI-augmented cooperative engineering assistant system.}
    \label{fig:architecture}
\end{figure}

As shown in Fig.~\ref{fig:architecture}, the architecture consists of a client side and a server side. The client side provides multiple user-facing interaction modalities, including a chat interface, a voice interface, and an image viewer. These interfaces are connected through a dispatcher, which manages communication between the user and the server. The chat interface supports text-based requests such as asking for task clarification, requesting a summary of recent design decisions, or checking the status of a requirement. The voice interface is intended for low-friction interaction during hands-on engineering work, allowing team members to report progress, raise problems, or capture design rationale without interrupting their current activity. The image viewer supports visual interaction with engineering artifacts, such as screenshots, diagrams, design outputs, or generated images.

On the server side, user identity and access are managed through a user credential server. This component separates user identification from task execution and enables the system to associate requests, updates, and permissions with individual users or team roles. Communication between the client and the server can occur through HTTP for standard request--response interactions and through WebSocket connections for continuous or interactive exchanges. This is important for cooperative workflows in which users may need ongoing updates, streaming responses, or asynchronous notifications about task status, requirement changes, or integration risks. The service selector is the central coordination component of the architecture. It receives user requests from the client dispatcher, interprets the intended action, and routes the request to the most appropriate specialized service. This design reflects the survey finding that ERC teams require support across multiple workflow dimensions rather than a single general-purpose assistant. For example, one service may handle task reporting and task operations, another may support requirement decomposition and compliance checking, while another may perform image generation, visual analysis, or interaction with design artifacts. The service selector therefore enables modularity: new services can be added without redesigning the entire system, and requests can be routed according to the current engineering context, user role, and available tools.

The specialized services interact with external tools and data sources, such as databases, design generators, version-control systems, visual-language models, or other engineering applications. This tool-integration layer is essential for moving beyond isolated conversational AI. For instance, a requirement-tracking service could extract ERC rules, decompose them into subsystem-level responsibilities, and connect them to tasks. A task-monitoring service could infer progress from communication logs or version-control activity and identify missing updates. A documentation service could capture design rationale from voice or chat interactions and update shared knowledge repositories. An integration-support service could analyze dependencies between subsystems, identify inconsistent assumptions, and notify relevant team members before late-stage conflicts appear during testing. The architecture addresses the main design requirements derived from the survey. First, requirement and compliance traceability are supported by services that connect external competition rules to internal tasks, responsibilities, and subsystem artifacts. Second, task clarification and monitoring are supported through services that decompose vague tasks, detect missing acceptance criteria, and infer progress from distributed activity. Third, knowledge transfer is supported through continuous capture of design rationale, decisions, and explanations from chat, voice, and engineering artifacts. Fourth, fragmented communication is addressed through cross-channel summarization and contradiction detection. Fifth, integration-risk detection is supported by services that analyze subsystem dependencies, interface assumptions, and design histories. In this way, the architecture operationalizes the empirical findings as a cooperative information system for hybrid human--AI engineering teams.

An illustrative workflow demonstrates the intended use of the architecture. A team member working on the rover's electrical subsystem notices a possible incompatibility between a wiring change and a mechanical interface. Instead of waiting for the next meeting, the member reports the issue through the voice interface. The dispatcher forwards the request to the server, where the service selector identifies that the issue concerns integration risk and routes it to the relevant service. The service retrieves related requirements, recent task updates, and available design artifacts through connected tools. It then summarizes the potential conflict, identifies affected subsystems, and suggests follow-up actions, such as notifying the mechanical group lead, checking the relevant requirement, or creating a validation task. The response is returned to the user and can also be logged as part of the shared project memory. This example illustrates how the architecture can reduce coordination overhead while preserving the autonomy of team members and the flexibility of existing ERC workflows.

\section{Discussion and Conclusions}

The findings of this study reveal that many of the challenges experienced by ERC teams are not primarily technical, but organizational and infrastructural in nature. Informal task allocation, unclear requirements, fragmented communication channels, limited documentation, and weak feedback mechanisms create conditions of sustained high demand with insufficient structural support. According to work-design and job demands--resources perspectives, such configurations can produce cognitive overload, uneven workload distribution, and increased reliance on late-stage corrective effort \cite{bakker2023job}. These effects are reflected in the high levels of rework, integration delays, and documentation debt reported by the teams.

A particularly important insight concerns the invisibility of coordination work. Much of the effort required to keep projects moving clarifying tasks, tracking dependencies, reminding others, reconstructing missing knowledge, and interpreting requirement changes remains implicit and unrecorded. Organizational research shows that such invisible labor often falls on a small subset of highly engaged individuals, increasing the risk of overload while masking systemic inefficiencies \cite{ilgen2005teams}. The ERC context amplifies this effect: time pressure, volunteer participation, high turnover, and academic constraints leave little room for heavy process enforcement. As a result, coordination problems are often addressed through informal effort rather than through stable information-system support.

The survey also shows that AI tools are already widely used by ERC participants, but mainly for isolated technical tasks such as coding, debugging, documentation drafting, or information lookup. While these uses can improve individual productivity, they do not directly address the cooperative workflow problems identified in this study. Supporting isolated tasks is insufficient when the main sources of inefficiency originate in task ambiguity, fragmented communication, weak requirement traceability, delayed feedback, and late discovery of subsystem incompatibilities. This distinction is central to the argument of the paper: the most valuable role of AI in such settings is not only to accelerate individual work, but to strengthen the cooperative information infrastructure through which engineering teams coordinate their work.

The proposed architecture responds to this need by positioning AI as an intermediary layer between team members, specialized services, and external engineering tools. Its client-side interfaces support low-friction interaction through chat, voice, and visual inputs, while the server-side service selector routes requests to task, requirement, documentation, communication, or integration-support services. In this way, the architecture operationalizes the requirements derived from the survey: requirement and compliance traceability, task clarification, progress awareness, continuous knowledge capture, communication summarization, and integration-risk detection. Rather than imposing a rigid project-management process, the architecture is intended to augment existing ERC workflows by making distributed information more visible, actionable, and reusable.

Voice-based interaction remains especially relevant in this context. Many ERC activities occur during hands-on design, assembly, testing, and debugging, where manual reporting or tool switching can be disruptive. A vocal AI assistant could allow team members to externalize status updates, encountered problems, emerging risks, or design rationale while continuing their work. For team leaders and group leaders, aggregated outputs from such interactions could provide a more continuous view of workload distribution, unresolved dependencies, and emerging bottlenecks. This form of support aligns with the broader principle that effective workload management depends less on tighter control and more on timely feedback, shared awareness, and adaptive coordination.

In conclusion, this paper argues that the greatest potential for AI-augmented engineering workflows lies not only in accelerating isolated technical tasks, but in reducing organizational friction and hidden coordination work. Based on a role-adaptive survey of ERC teams, we identified recurring bottlenecks in knowledge transfer, task assignment, requirement tracking, communication, rework, and integration. These findings were translated into requirements for AI-augmented cooperative engineering workflows and used to motivate an initial assistant-system architecture. The proposed architecture provides a foundation for connecting human-facing interfaces, service selection, specialized AI services, and engineering tools in support of hybrid human--AI engineering teams.

The study has limitations. The empirical findings are based on self-reported survey responses from ERC teams and therefore reflect perceived workflow challenges rather than direct observation of team behavior. In addition, the proposed architecture is an initial design and has not yet been evaluated in a deployed ERC setting. Future work will therefore focus on implementing selected services, integrating them with existing team tools, and empirically evaluating their effects on coordination quality, workload awareness, requirement traceability, integration-risk detection, and long-term sustainability of team performance.

\begin{credits}
\subsubsection{\ackname} 
The authors would like to thank the organizers of the ERC, in particular the European Space Foundation led by Łukasz Wilczyński, for their support and for enabling access to the participating teams. We also extend our sincere appreciation to the members, group leads, and team leads of the ERC 2025 teams who contributed their time and insights by responding to the survey: 4Space (Spain), AAU Space Robotics (Denmark), AGH Space Systems (Poland), ASU ROAR (Egypt), DJS Antariksh (India), FHNW Rover Team (Switzerland), FRoST (Germany), KNR Rover Team (Poland), Mars Rover Manipal (India), Orion Team (Poland), OzU Rover Team (Turkey), ProjectRED (Italy), Sapienza Technology Team (Italy), UPC Space Program (Spain), WARRYZN Space Robotics (Germany). This study would not have been possible without their willingness to share their experiences and challenges.
\end{credits}

\bibliographystyle{splncs04}
\bibliography{main}

\end{document}